\documentclass{ifacconf}

\usepackage{algorithm}
\usepackage{algcompatible} 
\usepackage{amsmath, amssymb}
\usepackage{caption} 
\usepackage[dvipsnames]{xcolor} 
\usepackage{graphicx} 
\usepackage{url}
\usepackage{booktabs}
\usepackage{subcaption}
\usepackage[numbers,sort&compress]{natbib}        
\usepackage{tikz}
\usetikzlibrary{positioning, calc, arrows.meta}

\newcommand{\C}{{\mathcal{C}}}
\newcommand{\R}{\mathbb{R}}

\newcommand{\T}{\mathcal{T}}
\newcommand{\V}{\mathcal{V}}
\newcommand{\E}{\mathcal{E}}
\begin{document}
\begin{frontmatter}

\title{Global Path Planner with Multi-Model Switching} 

\thanks[footnoteinfo]{This research was funded by the European Union's Horizon Europe under grant no. 101079342 (Fostering Opportunities Towards Slovak Excellence in Advanced Control for Smart Industries), and funded by the Italian Ministry of Education and Research in the framework of the FISA 2023 OCCAM Project and the ``FoReLab'' (Future-oriented Research Lab) Project (Departments of Excellence).
Pietro Gori and Francesco Iotti equally contributed.
© 2026 the authors. This work has been accepted to IFAC for
publication under a Creative Commons Licence CC-BY-NC-ND.}

\author[dii_cp]{Pietro Gori} 
\author[dii_cp]{Francesco Iotti} 
\author[stu]{Eduard Zelenay}
\author[panza]{Rastislav Marko}
\author[dii_cp]{Michele Pierallini}
\author[dii_cp]{Franco Angelini}
\author[destec]{Gabriele Pannocchia}
\author[dii_cp]{Manolo Garabini}

\address[dii_cp]{Centro di Ricerca ``Enrico Piaggio'' and the Dipartimento di Ingegneria dell'Informazione,  Universit\`a di Pisa, Largo Lucio Lazzarino 1, 56126 Pisa, Italy}
\address[panza]{Panza Robotics, National Center of Robotics, Ilkovicova 3, 841 04 Bratislava, Slovakia}
\address[stu]{Institute of Robotics and Cybernetics, Faculty of Electrical Engineering and Information
Technology, Slovak University of Technology in Bratislava, Ilkovicova 3, 812 19
Bratislava, Slovakia }
\address[destec]{Dipartimento di Ingegneria Civile e Industriale,  Universit\`a di Pisa, Largo Lucio Lazzarino 1, 56126 Pisa, Italy}

\begin{abstract}                
This work enhances global path planning via a pure‑pursuit controller with multi‑model kinematic switching that sustains plan fidelity across diverse terrains. The system includes a traversability graph for terrain analysis, a
Heading-Aware A* algorithm generating feasible paths, and a multi-model Pure
Pursuit controller for dynamic tracking. A core innovation is adaptive kinematic
modeling, enabling real-time switching between kinematic models based on terrain
features and robot states. This adaptability optimizes path efficiency and energy
use in challenging scenarios. We validate the approach in simulation on different platforms, namely, Artaban quadruped and X3 quadrotor drone, showcasing improved performance, robustness, and adaptability over standard baselines.
\end{abstract}


\end{frontmatter}

\section{Introduction}
\label{sec:introduction}

\vspace{-2mm}
Nowadays, mobile robots possess remarkable capabilities to traverse diverse terrains and navigate complex environments \citep{Hoeller2024}. However, despite their impressive mobility, these robots typically rely on human operators to determine their actions and behaviors \citep{iotti2025omniquad}. This dependence on external control introduces rigid constraints, significantly limiting the robot's ability to function effectively in scenarios where human intervention is impractical or potentially hazardous \citep{CHEN2025}.

Enhancing robot autonomy offers several tangible benefits. Autonomous platforms can continue operations in hazardous or communication-restricted environments, such as disaster zones, where teleoperation is challenging \citep{LOGANATHAN2023}. Furthermore, onboard decision-making enables robots to dynamically adjust their behaviors in real-time, greatly improving operational efficiency \citep{Wu2023}.

Path planning algorithms enhance robotic autonomy by enabling efficient navigation in complex, dynamic environments, improving reliability, safety, and energy efficiency through collision avoidance and path optimization. Their effectiveness hinges on balancing computational efficiency with real-time adaptability. Recent advances include application-specific approaches: Hirayama et al. \citep{HIRAYAMA2019} developed topological modeling for autonomous bulldozers in mining areas, integrating material handling and dynamics; Xing et al. \citep{XING2022} combined Seeker Optimization (SOA) for global paths with Actor-Critic reinforcement learning for local adaptability.

Path planning for quadruped robots operating in complex and unstructured environments requires robust methods that effectively combine global and local navigation strategies. In \citep{Jian2021}, the Coupling Two-Stage Path Planning (CTSP) approach integrates iterative global optimization with refined local cost functions, facilitating smooth transitions between global routes and local maneuvers. Similarly, sampling-based global planners combined with dynamics-aware local strategies explicitly incorporate robot kinematics and dynamics, enhancing traversal of challenging terrains \citep{Norby2020}.

Recent advancements focus on integrating perception and terrain analysis to improve real-time navigation. GPU-accelerated elevation mapping significantly reduces computational overhead by efficiently representing terrain geometry \citep{Miki2022}. In \citep{Dixit2025}, the STEP framework transforms noisy LiDAR data into risk-aware planning through Conditional Value-at-Risk (CVaR) metrics. TRG-Planner employs a direction-aware risk graph encoding terrain stability and reachability, enabling hierarchical real-time decision-making \citep{Lee2025TRG}. Additionally, machine learning techniques for traversability assessment and foothold scoring have proven effective in dynamic, uncertain environments \citep{Wellhausen2023}. The TAO framework incorporates both apparent and relative traversability metrics, further enhancing adaptive planning and motion control for effective hazard avoidance in mountainous terrains \citep{Yoo2024}.



Improving robot autonomy requires enabling robots to reason about and select the most suitable motion model for a given task. Existing approaches typically assume a single, fixed kinematic model, which limits robustness in diverse environments
\citep{Dasanayake2025}. 
In this work, we introduce a global path planning framework with multi-model switching, allowing the robot to autonomously switch between models based on hardware constraints, task requirements, and environmental conditions. 
The main contributions are:
\begin{itemize}
\vspace{-2mm}
    \item A novel Heading-Aware A* algorithm that incorporates orientation costs
into path planning, enabling smoother and more efficient trajectories.
    \item An adaptive trajectory follower capable of switching between kinematic models
according to the robot's internal state and environment.
    \item  A comprehensive validation on different robots in both simulation and
real-world experiments, showcasing robust and versatile performance.
\end{itemize}
\vspace{-2mm}

\section{Path Planning Algorithm}
\label{sec:path_planning}
\vspace{-2mm}
In this section, we introduce the path planning algorithm used to navigate the robot through the environment. 
The algorithm is designed to find an optimal path from a start point to a goal point while avoiding obstacles, ensuring efficiency and safety in its motion.
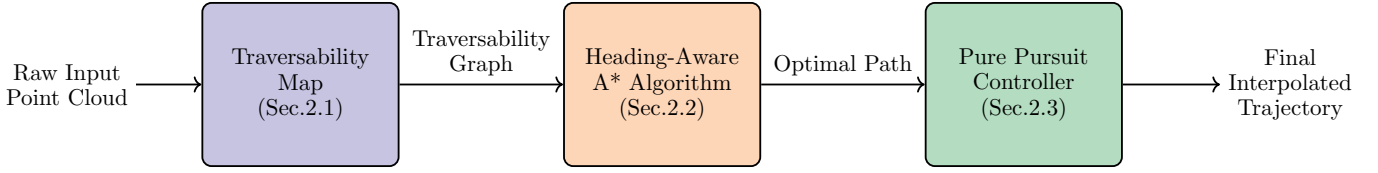
\begin{figure*}[ht]
    \centering
    \resizebox{\linewidth}{!}{
        \begin{tikzpicture}[
            node distance=2cm and 2.5cm,
            box/.style={rectangle, rounded corners, draw=black, thick, minimum width=3.0cm, minimum height=2.5cm, align=center},
            input/.style={above left, align=center},
            arrow/.style={->, thick}
        ]
        \node[box, fill=Blue!20,align=center] (traversability) {Traversability\\Map\\
        (Sec.\ref{subsec:traversability_map})
        };
        \node[box, fill=Orange!30, right=of traversability, align=center] (optim) {Heading-Aware\\A* Algorithm\\
        (Sec.\ref{subsec:optim_path})
        };
        \node[box, fill=Green!30, right=of optim,align=center] (controller) {Pure Pursuit\\Controller\\
        (Sec.\ref{subsec:pure_pursuit_controller})
        };

        \node[input, left=of traversability,align=center, xshift=1.5cm] (inputcloud) {Raw Input\\Point Cloud};

        \draw[arrow] (inputcloud) -- (traversability);
        \draw[arrow] (traversability) -- node[above,align=center] {Traversability\\Graph} (optim);
        \draw[arrow] (optim) -- node[above] {Optimal Path} (controller);
        \draw[arrow] (controller) -- ++(3.0,0) node[right,align=center] {Final\\Interpolated\\Trajectory};

        \end{tikzpicture}
    }
    \caption{Path planning pipeline using traversability computations, A* search, and Pure Pursuit control.}
    \label{fig:path_planning_pipeline}
\end{figure*}
The process begins with a raw point cloud of the environment as input, which is processed to compute traversability
information. This information is then used to build the traversability graph, as described in Section \ref{subsec:traversability_map}.
The Heading-Aware A* Algorithm, detailed in Section \ref{subsec:optim_path}, is employed to find the optimal path from the start point to the goal point.

Finally, the Pure Pursuit controller, discussed in Section \ref{subsec:pure_pursuit_controller}, refines the path,
generating a smooth trajectory for the robot to navigate through the environment.
Figure \ref{fig:path_planning_pipeline} illustrates the overall pipeline of the
path planning process.

\vspace{-2mm}
\subsection{Traversability Map}
\label{subsec:traversability_map}
\vspace{-2mm}
Initially, we obtain raw point cloud data of robot positions from sensor measurements. 
From these data, we identify collision-free areas based on terrain features such as height filtering, ground segmentation using RANSAC, and surface normal estimation. 
Points are classified according to surface normals: vertical normals indicate navigable regions, horizontal normals represent obstacles, and intermediate cases are assessed using grid-based slope and height evaluations. 
Navigable regions are further refined through coarser grid processing and validated with 2.5D maps. Finally, obstacles are grouped via spatial clustering (DBSCAN \citep{deng2020dbscan}), and nearby navigable points are reclassified to mitigate noise and potential misclassification.
We define $\C \in \R^{3}$ as the configuration space of the robot, representing all possible positions in the environment.
After this process, we identify two subsets:
\vspace{-2mm}
\begin{itemize}
    \item $\C_{\mathrm{free}} \subseteq \C$: the set of all traversable positions, corresponding to the navigable regions;
    \item $\C_{\mathrm{obs}} \subseteq \C$: the set of all non-traversable positions, corresponding to the obstacle regions.
\end{itemize}
\vspace{-2mm}
\begin{figure}[hb]
    \centering
    \includegraphics[width=0.6\linewidth]{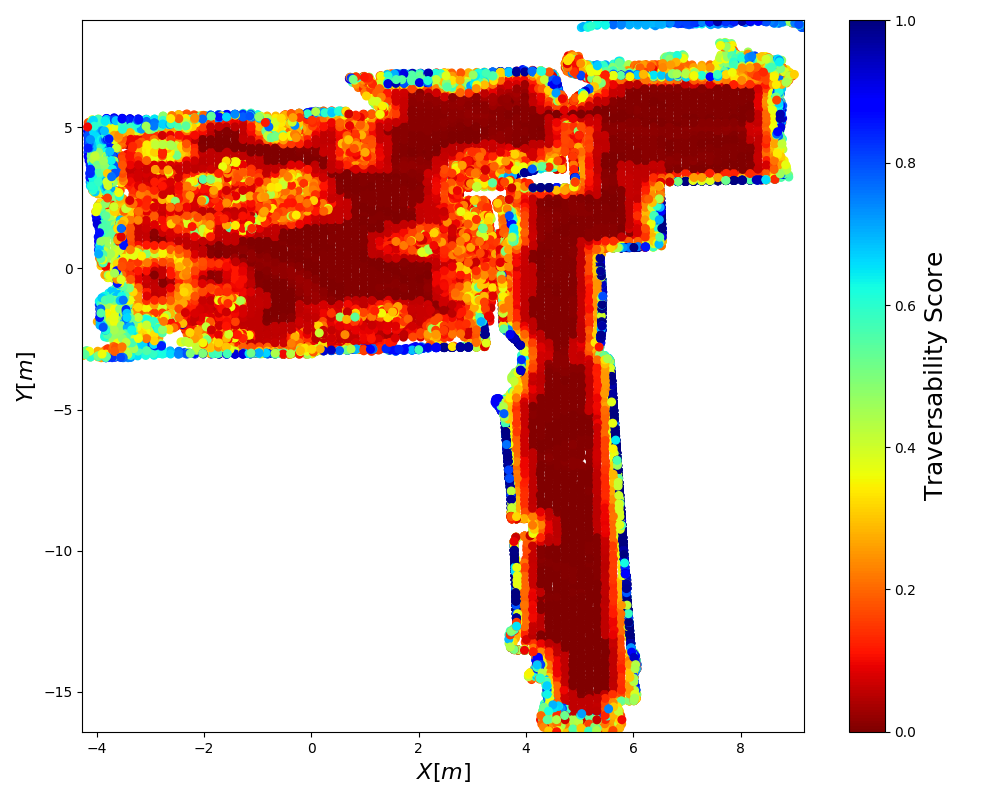}
    \caption{Example traversability map derived from raw point cloud data. 
    Navigable regions ($\C_{\mathrm{free}}$) and obstacle regions ($\C_{\mathrm{obs}}$) are visualized, 
    with color intensity indicating the local traversability cost: higher intensity corresponds to higher traversal difficulty or risk.}
    \label{fig:traversability_map}
\end{figure}
%
Given the sets $\C_{\mathrm{free}}$ and $\C_{\mathrm{obs}}$, we build the traversability
graph $\T_{G}= (\V, \E)$, which serves as the discrete representation of the
environment for path planning. Each vertex $v_{i} = (\mathbf{p}_{i}, c(v_{i}), \E_{i}) \in \V$ 
encodes a feasible robot position, 
where $\mathbf{p}_{i}\in \C_{\mathrm{free}}$ denotes the spatial position, $c(v_{i})$ is
the traversability cost, and $\E_{i} \subseteq \E$
is the set of incident edges of vertex $i$.
Edges $\xi_{i}^{j} \in \E_{i}$ connect pairs of vertices if a direct, collision-free transition between $v_{i}$ and $v_{j}$ is possible, determined by spatial proximity and obstacle clearance. 
This graph structure enables efficient search for optimal paths, leveraging both geometric connectivity and traversability information.

The cost function $c: \V \to \R$ for each vertex $v_i \in \V$ is computed as follows:
\begin{equation}
    \label{eq:node_cost}
    c(v_i) = w_{\mathrm{coll}} c_{\mathrm{coll}}(v_i) + w_{\mathrm{dist}} c_{\mathrm{dist}}(v_i) + w_{\mathrm{slope}} c_{\mathrm{slope}}(v_i),
\end{equation}
where $c_{\mathrm{coll}}, \,  c_{\mathrm{dist}}, c_{\mathrm{slope}} : \V \to \R$ are the collision, distance, and slope cost functions, respectively, 
and $w_{\mathrm{coll}}, \, w_{\mathrm{dist}}, w_{\mathrm{slope}}$ are the corresponding weights.
Figure \ref{fig:traversability_map} illustrates the cost evolution in the traversability map.
\subsubsection{Collision Cost}
The collision cost $c_{\mathrm{coll}}(v_i)$ penalizes a vertex $v_i\in\C_{\mathrm{free}}$ by its proximity to $\C_{\mathrm{obs}}$. 
Around $\mathbf{p}_i$ we consider a disk of radius 
$r_{\mathrm{robot}}=\sqrt{(L_{\mathrm{robot}}/2)^2+(W_{\mathrm{robot}}/2)^2}$, 
the robot’s XY circumscribed radius, where $L_{\mathrm{robot}}$ and $W_{\mathrm{robot}}$ are the robot's length and width, respectively. 
If any obstacle point lies inside this disk, we mark a hard collision; otherwise we apply a soft penalty 
$P(dist)~= A~\cdot \exp\left(-\left(\frac{dist}{\sigma}\right)^2\right)$,
where $\mathrm{dist}$ is the minimum Euclidean distance to $\C_{\mathrm{obs}}$, $A$ is the maximum penalty, and $\sigma$ controls the decay rate.. 
This encourages clearance while remaining permissive in narrow passages.

\subsubsection{Distance Cost}
The distance cost function $c_{\mathrm{dist}}(v_i)$ is defined as the Euclidean distance from the vertex position $\mathbf{p}_{i}$ 
to the next vertex position $\mathbf{p}_j$ in the traversability graph.
This cost is computed as:
\begin{equation}
    \label{eq:distance_cost}
    c_{\mathrm{dist}}(v_i) = \| \mathbf{p}_j - \mathbf{p}_i \|_2,
\end{equation}
where $\| \cdot \|_2$ denotes the Euclidean norm.
\subsubsection{Slope Cost}
The slope cost function $c_{\mathrm{slope}}(v_{i})$ measures how steep the terrain is around a vertex $v_{i}$,
since steeper slopes are more difficult to traverse.
To compute it, neighboring points are first translated so that $v_{i}$ lies at the origin $(0,0,0)$.
A plane of the form $z = a x + b y + c$ is then fitted to these points using least squares. 
Because the plane passes through the origin, $c \approx 0$.
The plane's normal vector is $\mathbf{n} = [-a, -b, 1]$, and the local slope angle is the angle between
$\mathbf{n}$ and the vertical axis $\mathbf{\hat{k}} = [0,0,1]$:
\begin{equation}
    \label{eq:slope_angle}
    \theta = \arccos\!\left(\frac{\mathbf{n} \cdot \hat{\mathbf{k}}}{\|\mathbf{n}\|\|\hat{\mathbf{k}}\|}\right)
= \arccos\!\left(\frac{1}{\sqrt{a^{2}+b^{2}+1}}\right).
\end{equation}
Finally, $c_{\mathrm{slope}}(v_{i})$ is defined as a function of $\theta$, with
larger angles corresponding to higher traversal cost.
%
%
\vspace{-2mm}
\subsection{Heading-Aware A* Algorithm}
\label{subsec:optim_path}
\vspace{-2mm}
We compute optimal paths from a start vertex, i.e., $v_{\textnormal{start}}=(\mathbf{p}_{\textnormal{start}}, c(v_{\textnormal{start}}), \E_{\textnormal{start}})$, to a goal vertex, i.e., $v_{\textnormal{goal}}=(\mathbf{p}_{\textnormal{goal}}, c(v_{\textnormal{goal}}), \E_{\textnormal{goal}})$, on a traversability graph $\T_{G}= (\V, \E)$ using a heading-aware A* variant. 
Each search state combines vertex $v_{i}\in \V$ and arrival heading $\psi_{i}$ as $(v_{i}, \psi_{i})$, enabling orientation-dependent cost penalties (e.g., for sharp turns and terrain slope) that generate smooth and feasible trajectories also for non-holonomic robots.
The complete procedure is detailed in Algorithm~\ref{alg:modified_astar}.
\begin{algorithm}[t]
    \caption{Heading-Aware A* Algorithm}
    \label{alg:modified_astar}
    \begin{algorithmic}[1]
        \REQUIRE Traversability graph $\mathcal{T}_{G}=(\mathcal{V},\mathcal{E})$, start $v_{s}$, goal $v_{g}$, headings $\psi_{s}$, $\psi_{g}$
        \ENSURE Optimal path from $v_{s}$ to $v_{g}$ with heading constraints
        \STATE Initialize queue with $(v_{s}, \psi_{s})$; $g[\cdot] \gets \infty$; $g[(v_{s}, \psi_{s})] \gets 0$
        \STATE $f[(v_{s}, \psi_{s})] \gets h(v_{s}, \psi_{s})$; $\text{parent}[\cdot] \gets \textsc{Null}$
        \WHILE{queue not empty}
            \STATE Pop $(v_{c}, \psi_{c})$ with $\min f$
            \IF{$v_{c} = v_{g}$} \textbf{return} path via $\text{parent}$ \ENDIF
            \FOR{each neighbor $v_{n}$ of $v_{c}$}
                \STATE $\mathbf{d} \gets \mathbf{p}_{n} - \mathbf{p}_{c}$; $\psi_{\text{edge}} \gets \mathrm{atan2}(d_{y}, d_{x})$
                \STATE $\Delta\psi \gets \textsc{AngleDiff}(\psi_{\text{edge}}, \psi_{c})$
                \STATE $C_{\text{step}} \gets c(v_{n}) + w_{\psi}|\Delta\psi| + w_{\text{slope}}\cdot \sin\theta \cdot \cos\phi$
                \STATE $g_{\text{new}} \gets g[(v_{c}, \psi_{c})] + C_{\text{step}}$
                \IF{$g_{\text{new}} < g[(v_{n}, \psi_{\text{edge}})]$}
                    \STATE $g[(v_{n}, \psi_{\text{edge}})] \gets g_{\text{new}}$
                    \STATE $f \gets g_{\text{new}} + \|\mathbf{p}_{n} - \mathbf{p}_{g}\|_2 + w_{\text{goal}}\cdot|\psi_{g} - \psi_{\text{edge}}|$
                    \STATE $\text{parent}[(v_{n}, \psi_{\text{edge}})] \gets (v_{c}, \psi_{c})$
                    \STATE Enqueue $(v_{n}, \psi_{\text{edge}})$
                \ENDIF
            \ENDFOR
        \ENDWHILE
    \end{algorithmic}
\end{algorithm}
The core of the algorithm resamble the core of the classic A* algorithm, maintaining a min-priority queue of candidate 
states ordered by their estimated total cost ($f$-score), a cumulative path cost map ($g$-score) for each visited state, 
and a predecessor tracking map for path reconstruction.
The cumulative cost, $g(v_{i}, \psi_{i})$, to reach a state $(v_{i}, \psi_{i})$ is determined by summing the costs 
of individual transitions. 
For each transition from a current state $(v_{j}, \psi_{j})$ to a successor state $(v_{i}, \psi_{i})$, the step cost, 
$C_{\textnormal{step}}$, consists of three primary components:
\begin{itemize}
    \item \textbf{Node Cost ($C_{\textnormal{base}}$):} An inherent cost associated with traversing the edge $\xi_{j}^{i}$, representing the baseline effort required for movement. 
    Specifically, this cost is defined by the node cost $c(v_{i})$, i.e., \eqref{eq:node_cost}, of the successor vertex $v_{i}$, which incorporates the information from the traversability graph.
    \item \textbf{Orientation Change Penalty ($C_{\textnormal{head}}$):} A penalty for deviations in the robot's heading. 
    The orientation of the traversed edge, $\psi_{\textnormal{edge}}$, is computed from the displacement vector between $\mathbf{p}_{j}$ and $\mathbf{p}_{i}$. 
    The absolute difference between this edge orientation and the robot's current orientation, $\Delta\psi = \psi_{\textnormal{edge}} - \psi_{j}$, 
    is normalized to $(-\pi, \pi]$ and weighted by a coefficient $w_{\psi}$:
    \begin{equation}
        \label{eq:orientation_change_penalty}
        C_{\textnormal{head}} = w_{\psi} \cdot |\Delta\psi|.
    \end{equation}
    \item \textbf{Terrain Interaction Cost ($C_{\textnormal{slope}}$):} A terrain-adaptive cost that penalizes movements based on the local slope
    and the travelling direction relative to the surface normal. 
    This cost is calculated as:
    \begin{equation}
        \label{eq:terrain_interaction_cost}
        C_{\textnormal{slope}} = w_{\mathrm{\textnormal{slope}}} \cdot \sin(\theta) \cdot \cos(\phi),
    \end{equation}
    where $w_{\mathrm{\textnormal{slope}}}$ is a weighting factor, $\theta$ is the slope angle at $v_{j}$ \eqref{eq:slope_angle},
    and $\phi$ is the angle between the normalized movement vector (from $\mathbf{p}_{j}$ to $\mathbf{p}_{i}$) and the surface normal at $v_{j}$,
    i.e., $\mathbf{n}$. 
    This formulation increases the cost for traversing steeper terrain, especially when moving uphill against the normal.
    However, it allows the robot to traverse steep slopes if necessary.
\end{itemize}
The total step cost for a transition is the sum of these components:
\begin{equation}
    \label{eq:total_step_cost}
    C_{\textnormal{step}}= C_{\textnormal{base}}+ C_{\textnormal{head}}+ C_{\textnormal{slope}}.
\end{equation}
The cumulative $g$-score for a successor state is updated if a newly discovered path
yields a lower total cost.

The heuristic function, $h(v_{i}, \psi_{i})$, provides an admissible estimate of
the cost-to-go from a successor state $(v_{i}, \psi_{i})$ to the goal by leveraging the Euclidean distance between the spatial coordinates of
$v_{i}$ and the target node $v_{\textnormal{goal}}$.
Additionally, if a specific goal orientation $\psi_{\textnormal{goal}}$ is provided, a weighted
penalty for the angular deviation between the successor's orientation $\psi_{s}$
and $\psi_{\textnormal{goal}}$ is incorporated into the heuristic:
\begin{equation}
    \label{eq:orientation_heuristic}
    h(v_{i}, \psi_{i}) = ||\mathbf{p}_{i}- \mathbf{p}_{\textnormal{goal}}||_{2} + w^{\textnormal{goal}}_{\psi} \cdot |\psi_{i} - \psi_{\textnormal{goal}}|,
\end{equation}
where $w^{\textnormal{goal}}_{\psi}$ is a weight factor that balances the importance of spatial
distance against the orientation penalty.
The $f$-score is then computed as the sum of the cumulative cost and the heuristic.

The heading-aware A* algorithm proceeds by repeatedly selecting the state with the lowest total
estimated cost ($f$-score) from the priority queue. 
Once the goal is reached, the optimal path is reconstructed by tracing back through the predecessor map from the goal to the start.
\vspace{-2mm}
\subsection{Multi-Model Pure Pursuit Controller}
\label{subsec:pure_pursuit_controller}
\vspace{-2mm}
The pure pursuit controller is a geometric path tracking method widely used for mobile robots \citep{samuel2016review, Lee2025TRG}. 
In our framework, it tracks the path generated by the Heading-Aware A* planner (Sec.~\ref{subsec:optim_path}) and computes velocity commands for both planar and non-planar motion.

At each control cycle, the controller selects a lookahead point at distance $L_d$ along the reference path and computes the tracking error in the robot frame. 
According to the path geometry and robot state, the controller dynamically switches between unicycle and omnidirectional kinematic models.

In the following, we briefly describe the two kinematic models used in our pure pursuit controller.

For planar motion, the unicycle model is:
\begin{align}
\dot{x} &= u^{\mathrm{uni}}\cos\psi, \\
\dot{y} &= u^{\mathrm{uni}}\sin\psi, \\
\dot{\psi} &= \omega .
\end{align}

For non-planar motion, the omnidirectional model is:
\begin{align}
\dot{x} &= u_x^{\mathrm{omni}}, \\
\dot{y} &= u_y^{\mathrm{omni}}, \\
\dot{z} &= u_z^{\mathrm{omni}} .
\end{align}
The model selection process is designed to ensure that the robot always follows the most efficient control strategy, adapting to both the geometry of the path and the robot's current state. 

\begin{figure}[t]
    \centering
    \includegraphics[width=0.6\linewidth, trim={8.5cm 2.5cm 8.5cm 4.0cm}, clip]{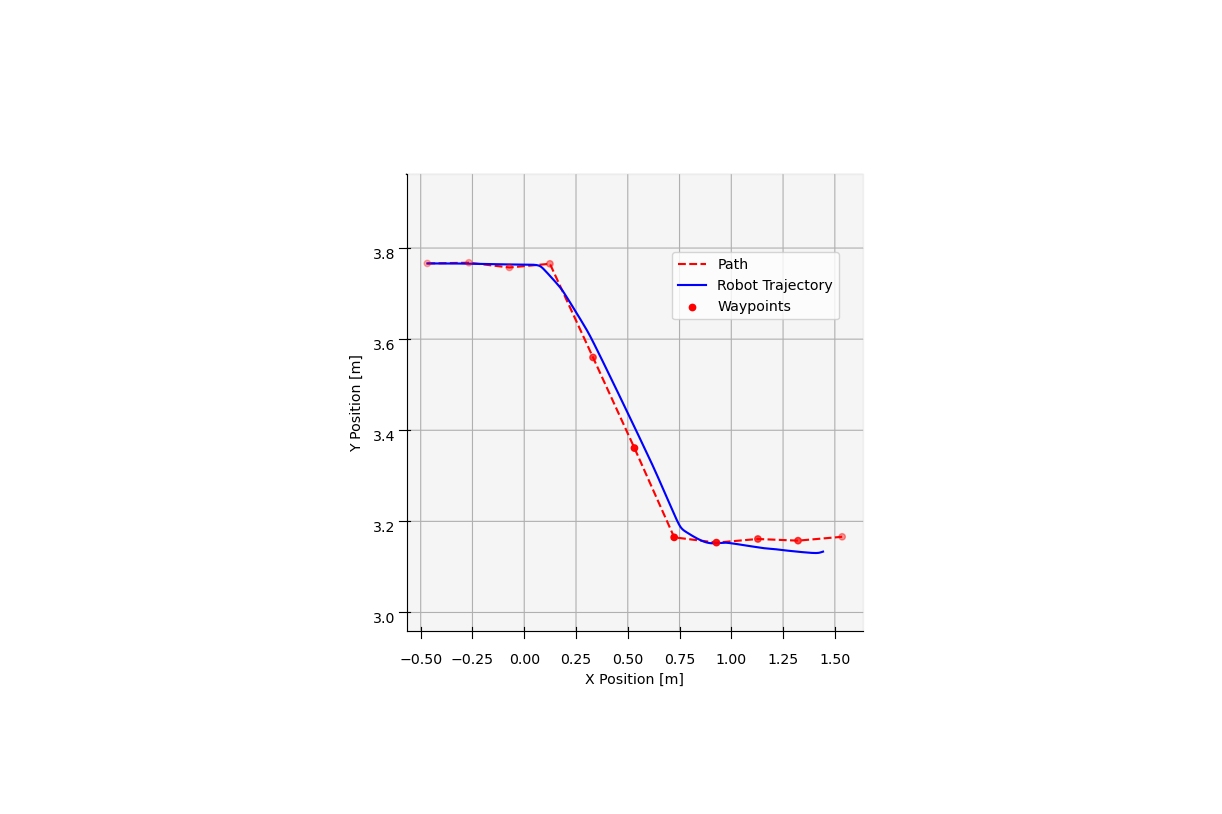}
    \caption{Top-down view of the planned trajectory in the XY plane.}
    \label{fig:planning_xy}
\end{figure}

The tracking error in the robot frame is defined as
\begin{align}
\mathbf{e}_{\mathcal{R}}=
\mathbf{R}_{\mathcal{RW}}^\top
\begin{bmatrix}
x_l-x_r\\
y_l-y_r\\
z_l-z_r
\end{bmatrix}.
\end{align}

From $\mathbf{e}_{\mathcal{R}}$, the planar distance $d$, heading error $\Delta\psi$, and vertical displacement $z_{\mathrm{disp}}$ are computed as
\begin{align}
d &= \sqrt{e_x^2+e_y^2}, \\
\Delta\psi &= \mathrm{atan2}(e_y,e_x), \\
z_{\mathrm{disp}} &= e_z .
\end{align}

The controller selects the unicycle model when
\begin{align}
|\Delta\psi|<\delta_\psi
\quad \text{and} \quad
|z_{\mathrm{disp}}|<\delta_z,
\end{align}
otherwise, the omnidirectional model is used.
Note that $\delta_\psi$ and $\delta_z$ are predefined thresholds.

After selecting the appropriate kinematic model, the controller computes the velocities required to reach the lookahead point.

\subsubsection{Omnidirectional Control}
In omnidirectional mode, the commanded velocity is proportional to the tracking error:
\begin{align}
\mathbf{u}^{\mathrm{omni}}=
\begin{cases}
\mathbf{0}, & D<\epsilon,\\
S\,\mathbf{e}_{\mathcal{R}}, & \text{otherwise},
\end{cases}
\end{align}
where
\begin{align}
S=\min\left(\frac{u_{\max}^{\mathrm{omni}}}{D},1\right),
\quad
D=\|\mathbf{e}_{\mathcal{R}}\|_2 .
\end{align}
\subsubsection{Unicycle Control}

In unicycle mode, the linear velocity is proportional to the distance error:
\begin{align}
u^{\mathrm{uni}} = k_v d \cos(\Delta\psi),
\end{align}
saturated to the admissible velocity range $[-u^{\mathrm{uni}}_{\mathrm{max}}, u^{\mathrm{uni}}_{\mathrm{max}}]$.

The angular velocity is computed from the heading error:
\begin{align}
\omega = k_\omega \Delta\psi,
\end{align}
where $k_v$ and $k_{\omega}$ are proportional gains for linear and angular velocity, respectively.
Fig.~\ref{fig:planning_xy} illustrates the resulting trajectory obtained with the proposed controller.

\section{Numerical Results}
\label{sec:numerical_results}
\vspace{-2mm}
The path planning method described in Sec. \ref{sec:path_planning} is then validated through both simulation and real-world experiments. 
The goal is to showcase the effectiveness of the proposed approach in generating feasible and efficient paths for robots in various environments.
\vspace{-2mm}
\begin{figure*}[htbp]
    \centering
    \includegraphics[
        width=0.95\textwidth]{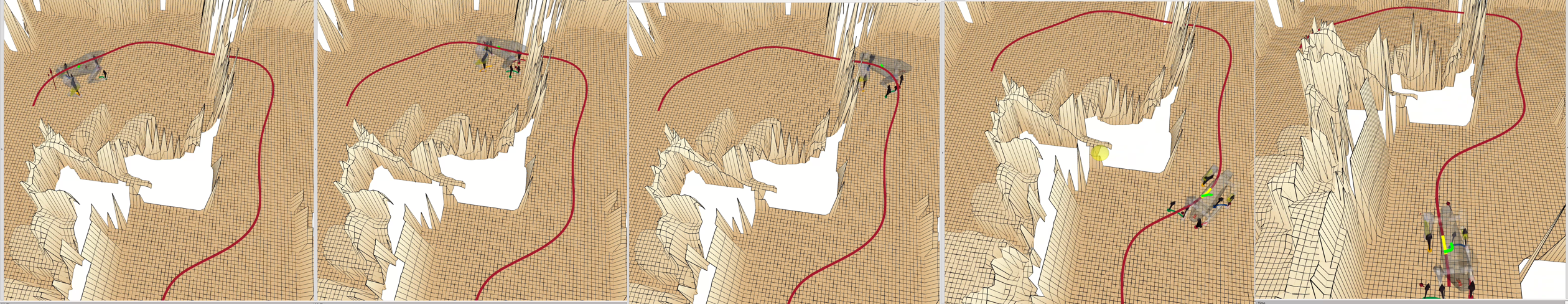}
    \vspace{-2mm}
    \caption{Trajectory executed by the Artaban robot during the simulation experiment.}
    \label{fig:robot_navigation}
    \vspace{-4mm}
\end{figure*}

\subsection{Simulation}
\label{subsec:simulation}
\vspace{-2mm}
The proposed path planner was validated using the Artaban robot \citep{vdurikovivc2024artaban}, developed by Panza Robotics\footnote{Company website: \url{[https://panzarobotics.com}}, and an X3 quadrotor drone.
\vspace{-2mm}
\subsubsection{Artaban quadruped}is a legged robot for hazardous environments, featuring cardan joints and four-bar linkages enabling multidirectional locomotion, with three actuators per leg. 
Its driver architecture employs PD controllers and inverse dynamics to handle unique kinematic loops in the rear legs, ensuring precise torque control and stability. The system compensates for friction and inertia, reducing sim-to-real discrepancies. 
The robot runs Quad-SDK \citep{norby2022quad}, an open-source ROS-based framework integrating planning, control, and estimation. The global planner node, implemented in C++, interfaces with Quad-SDK, defining the robot’s position and velocity to follow the trajectory.

\begin{figure}[t]
    \centering
    \includegraphics[width=0.65\linewidth, trim={2.2cm 1cm 3cm 1cm}, clip]{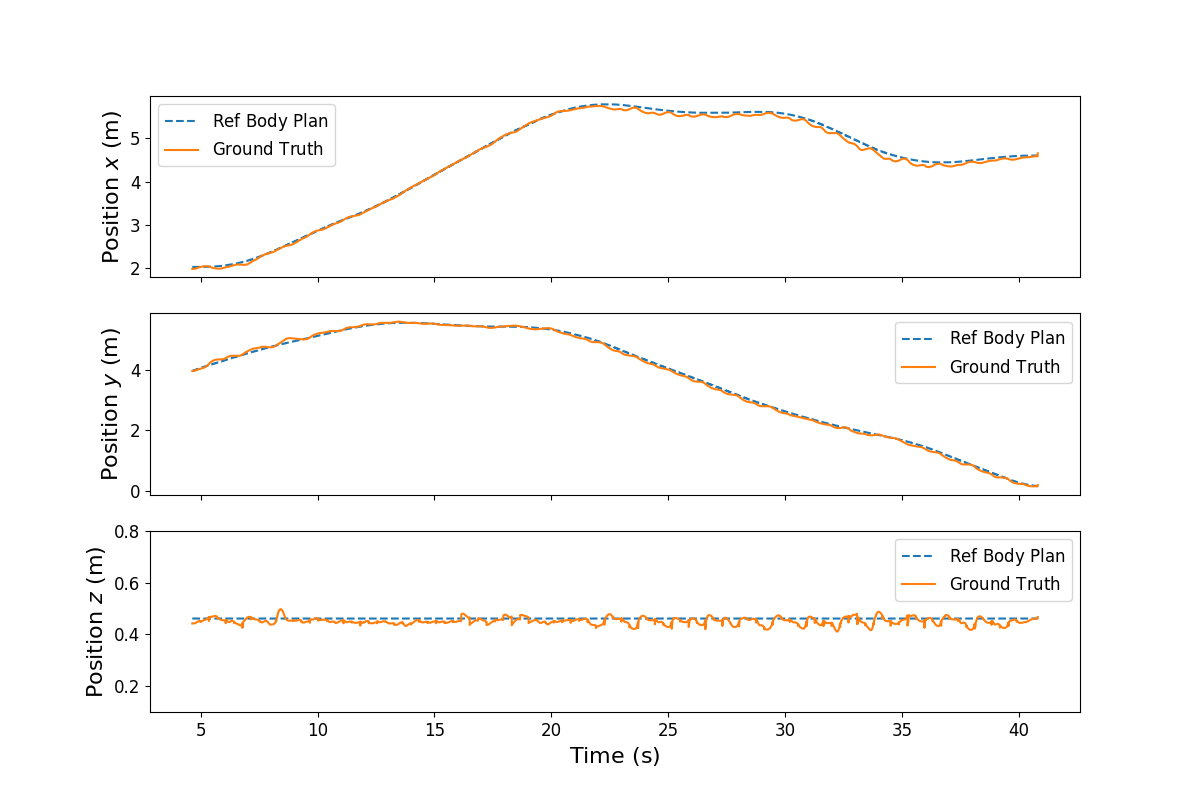}
    \caption{Body position tracking vs. reference in x, y, and z axes.}
    \label{fig:pos_sim}
    \vspace{-2mm}
\end{figure}

\begin{figure}[t]
    \centering
    \includegraphics[width=0.7\linewidth, trim={0cm 0cm 0cm 0cm}, clip]{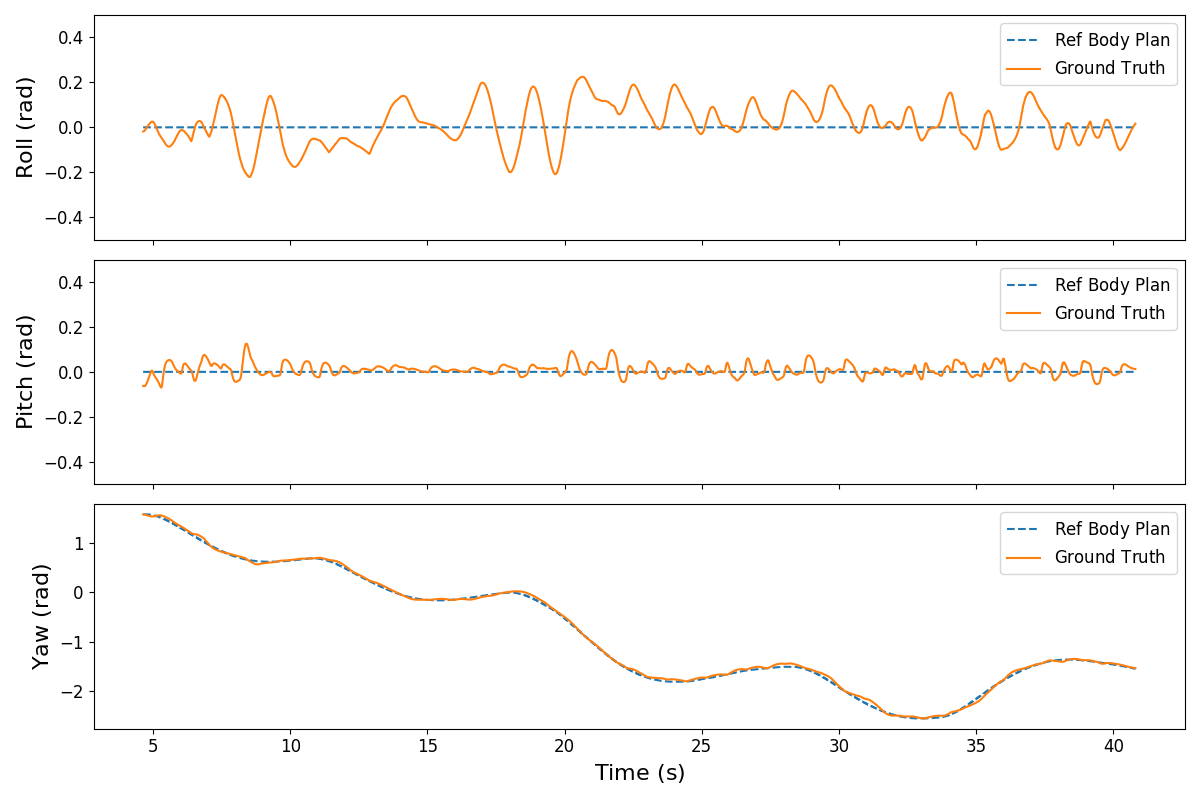}
    \caption{Body orientation tracking vs. reference}
    \label{fig:angle_sim}
\end{figure}

The validation scenario requires the robot to navigate through a narrow doorway before entering an open space area. 
During the door traversal, the robot adopts a unicycle-like motion, aligning its longitudinal axis with the doorway 
to ensure successful passage. 
Once through the door and into the open space, the robot transitions to more flexible, omnidirectional movements, 
leveraging its full kinematic capabilities to efficiently reach the goal. 
This showcases the planner's ability to adapt motion strategies based on environmental constraints.
Figures \ref{fig:pos_sim} and \ref{fig:angle_sim} report the simulation results.
The reference trajectories for position and orientation closely match the ground truth data, particularly in the horizontal plane (X and Y axes) and yaw orientation. Minor deviations are observed in the vertical axis (Z), where the algorithm generates a smooth, idealized trajectory compared to the robot’s ground truth, which exhibits realistic oscillations caused by physical dynamics and interactions with terrain, as depicted in Fig. \ref{fig:pos_sim}. Similarly, the planned orientations maintain minimal roll and pitch angles, reflecting simplified assumptions in the planner, whereas the simulated robot naturally experiences more pronounced body tilts, Fig. \ref{fig:angle_sim}.

\vspace{-2mm}
\subsubsection{X3 quadrotor drone}
\begin{figure}[t]
    \centering
    \includegraphics[width=0.7\linewidth, trim={1cm 0.5cm 0cm 0cm}, clip]{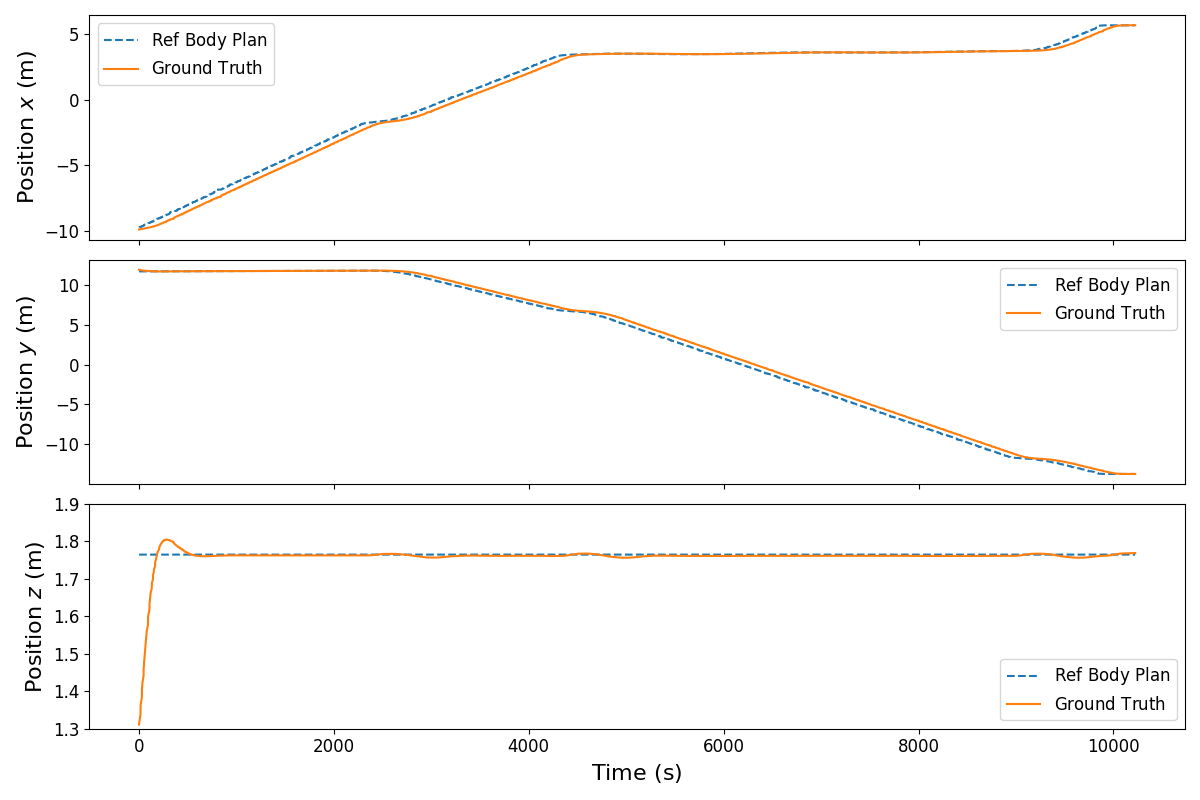}
    \caption{Body position tracking vs. reference in x, y, and z
axes.}
    \label{fig:drone}
\end{figure}

is a type of unmanned aerial vehicle (UAV) with four rotors arranged in a cross (X) configuration. It is equipped with a ROS-based flight controller capable of executing velocity commands.
In the drone simulation, our aim is to validate the planner's capability to generate smooth and efficient trajectories in a 3D space.
The drone navigates through a complex environment, requiring precise control over its position and orientation to avoid obstacles and reach the target location.
Despite the complexity of the environment, the planner successfully computes trajectories that the drone can follow accurately. Similarly to the simulation with the quadruped robot, Fig. \ref{fig:drone} shows a good tracking performance.
%
\vspace{-2mm}
\subsection{Real-World Experiment}
\label{subsec:real_world_experiment}
\begin{figure}[t]
    \centering
    \includegraphics[width=0.7\linewidth, trim={1.0cm 1cm 3cm 0.0cm}, clip]{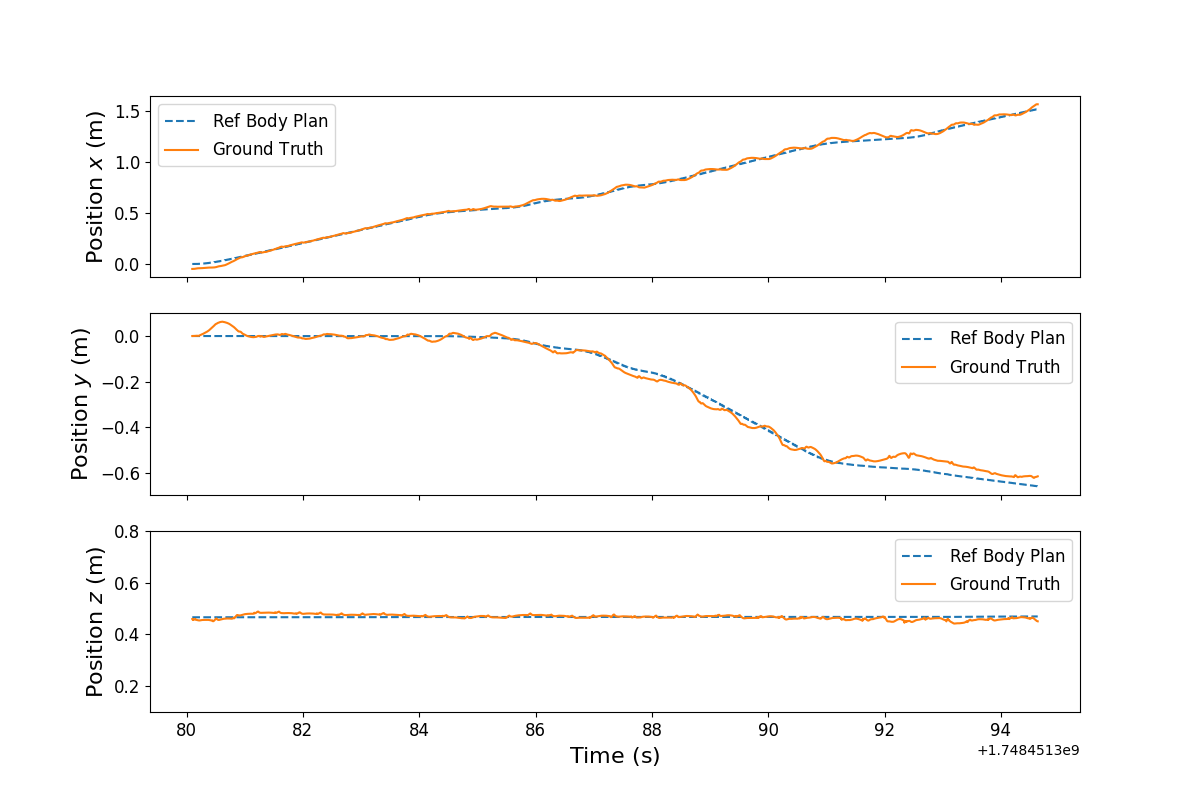}
    \caption{Body position tracking vs. reference }
    \label{fig:pos_real}
\end{figure}

\begin{figure}[t]
    \centering
    \includegraphics[width=0.7\linewidth, trim={0cm 0cm 0cm 0cm}, clip]{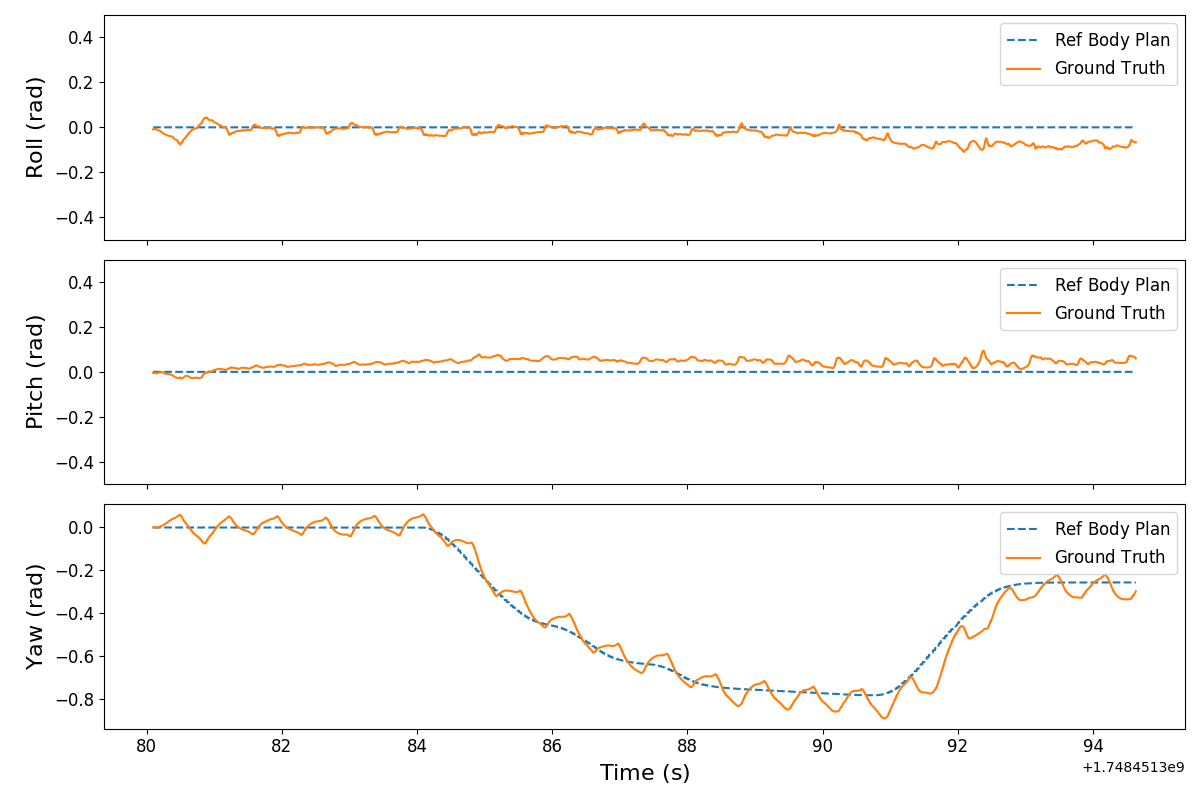}
    \caption{Body orientation tracking vs. reference}
    \label{fig:angle_real}
\end{figure}

\vspace{-2mm}
To further evaluate the robustness of our approach, we conducted experiments on the real Artaban hardware. 
The results demonstrate that the robot successfully completed the path, validating the effectiveness of the planner in a real-world setting. 
For the real robot scenario, the reference trajectory, Fig. \ref{fig:planning_xy}, computed by the planner also closely tracks the actual robot movement. 
The positional tracking, depicted in Fig.\ref{fig:pos_real}, is consistent with the simulation, with minor discrepancies attributed primarily to real-world sensor noise and execution imperfections. 
Orientation tracking, inferred to be similar to simulation results, further supports the algorithm's capability in effectively managing heading (yaw) and maintaining nominal body orientations (roll and pitch), although the dynamic influences in a real-world environment slightly amplify discrepancies observed in simulation, Fig.\ref{fig:angle_real}.

\section{Conclusion}
\vspace{-4mm}
This paper presented an integrated framework for robot navigation in complex planar and non-planar environments. 
The proposed approach combines a Heading-Aware A* planner operating on a traversability graph with a multi-model pure pursuit controller for adaptive trajectory tracking. 
The planner generates feasible and smooth paths by incorporating orientation information into the graph search process, while the controller dynamically switches between unicycle and omnidirectional motion models according to the local terrain and path geometry. 
The framework was validated in simulation and real-world experiments using the Artaban robot, demonstrating reliable navigation through narrow passages, elevation changes, and open environments. Future work will focus on dynamic obstacle avoidance and online replanning in more challenging scenarios.

\bibliography{Bibliography.bib}             

@inproceedings{deng2020dbscan,
  title={DBSCAN clustering algorithm based on density},
  author={Deng, Dingsheng},
  booktitle={2020 7th international forum on electrical engineering and automation (IFEEA)},
  pages={949--953},
  year={2020},
  organization={IEEE}
}

@article{samuel2016review,
  title={A review of some pure-pursuit based path tracking techniques for control of autonomous vehicle},
  author={Samuel, Moveh and Hussein, Mohamed and Mohamad, Maziah Binti},
  journal={International Journal of Computer Applications},
  volume={135},
  number={1},
  pages={35--38},
  year={2016},
  publisher={Foundation of Computer Science}
}

@ARTICLE{Lee2025TRG,
  author={Lee, Dongkyu and Nahrendra, I Made Aswin and Oh, Minho and Yu, Byeongho and Myung, Hyun},
  journal={IEEE Robotics and Automation Letters}, 
  title={TRG-Planner: Traversal Risk Graph-Based Path Planning in Unstructured Environments for Safe and Efficient Navigation}, 
  year={2025},
  volume={10},
  number={2},
  pages={1736-1743},
}

@ARTICLE{Jian2021,
  author={Jian, Zhiqiang and Zhang, Songyi and Chen, Shitao and Nan, Zhixiong and Zheng, Nanning},
  journal={IEEE Robotics and Automation Letters}, 
  title={A Global-Local Coupling Two-Stage Path Planning Method for Mobile Robots}, 
  year={2021},
  volume={6},
  number={3},
  pages={5349-5356},
  }

@INPROCEEDINGS{Miki2022,
  author={Miki, Takahiro and Wellhausen, Lorenz and Grandia, Ruben and Jenelten, Fabian and Homberger, Timon and Hutter, Marco},
  booktitle={2022 IEEE/RSJ International Conference on Intelligent Robots and Systems (IROS)}, 
  title={Elevation Mapping for Locomotion and Navigation using GPU}, 
  year={2022},
  volume={},
  number={},
  pages={2273-2280},
}

@INPROCEEDINGS{Norby2020,
  author={Norby, Joseph and Johnson, Aaron M.},
  booktitle={2020 IEEE/RSJ International Conference on Intelligent Robots and Systems (IROS)}, 
  title={Fast Global Motion Planning for Dynamic Legged Robots}, 
  year={2020},
  volume={},
  number={},
  pages={3829-3836},
 }

@ARTICLE{Dixit2025,
  author={Dixit, Anushri and Fan, David D. and Otsu, Kyohei and Dey, Sharmita and Agha-Mohammadi, Ali-Akbar and Burdick, Joel W.},
  journal={IEEE Transactions on Field Robotics}, 
  title={STEP: Stochastic Traversability Evaluation and Planning for Risk-Aware Navigation; Results From the DARPA Subterranean Challenge}, 
  year={2025},
  volume={2},
  number={},
  pages={81-99},
 }

@ARTICLE{Yoo2024,
  author={Yoo, Se-Wook and Son, E-In and Seo, Seung-Woo},
  journal={IEEE Robotics and Automation Letters}, 
  title={Traversability-Aware Adaptive Optimization for Path Planning and Control in Mountainous Terrain}, 
  year={2024},
  volume={9},
  number={6},
  pages={5078-5085},
}

@ARTICLE{Wellhausen2023,
  author={Wellhausen, Lorenz and Hutter, Marco},
  journal={Field Robotics}, 
  title={ArtPlanner: Robust Legged Robot Navigation in the Field}, 
  year={2023},
  volume={3},
  number={},
  pages={413-434},
 }

@article{HIRAYAMA2019,
title = {Path planning for autonomous bulldozers},
journal = {Mechatronics},
volume = {58},
pages = {20-38},
year = {2019},
issn = {0957-4158},
author = {Masami Hirayama and Jose Guivant and Jayantha Katupitiya and Mark Whitty},
}

@article{XING2022,
title = {Robot path planner based on deep reinforcement learning and the seeker optimization algorithm},
journal = {Mechatronics},
volume = {88},
pages = {102918},
year = {2022},
issn = {0957-4158},
author = {Xiangrui Xing and Hongwei Ding and Zhuguan Liang and Bo Li and Zhijun Yang},
}

@article{Hoeller2024,
author = {David Hoeller  and Nikita Rudin  and Dhionis Sako  and Marco Hutter },
title = {ANYmal parkour: Learning agile navigation for quadrupedal robots},
journal = {Science Robotics},
volume = {9},
number = {88},
pages = {eadi7566},
year = {2024},
}

@article{CHEN2025,
title = {A survey of autonomous robots and multi-robot navigation: Perception, planning and collaboration},
journal = {Biomimetic Intelligence and Robotics},
volume = {5},
number = {2},
pages = {100203},
year = {2025},
issn = {2667-3797},
author = {Weinan Chen and Wenzheng Chi and Sehua Ji and Hanjing Ye and Jie Liu and Yunjie Jia and Jiajie Yu and Jiyu Cheng},
}

@inproceedings{vdurikovivc2024artaban,
  title={Artaban: Adaptive Control for Multidirectional Locomotion},
  author={{\v{D}}urikovi{\v{c}}, Roman and Ma{\v{c}}icov{\'a}, Zuzana},
  booktitle={2024 New Trends in Signal Processing (NTSP)},
  pages={1--8},
  year={2024},
  organization={IEEE}
}

@inproceedings{norby2022quad,
  title={Quad-SDK: Full stack software framework for agile quadrupedal locomotion},
  author={Norby, Joseph and Yang, Yanhao and Tajbakhsh, Ardalan and Ren, Jiming and Yim, Justin K and Stutt, Alexandra and Yu, Qishun and Flowers, Nikolai and Johnson, Aaron M},
  booktitle={ICRA Workshop on Legged Robots},
  pages={1--5},
  year={2022},
  organization={sn}
}

@article{LOGANATHAN2023,
title = {A systematic review on recent advances in autonomous mobile robot navigation},
journal = {Engineering Science and Technology, an International Journal},
volume = {40},
pages = {101343},
year = {2023},
issn = {2215-0986},
author = {Anbalagan Loganathan and Nur Syazreen Ahmad},
}

@article{Wu2023,
    author = {Wu, Mingyu and Yeong, Che Fai and Su, Eileen Lee Ming and Holderbaum, William and Yang, Chenguang},
    title = {A review on energy efficiency in autonomous mobile robots},
    journal = {Robotic Intelligence and Automation},
    volume = {43},
    number = {6},
    pages = {648-668},
    year = {2023},
    month = {09},
    issn = {2754-6969},
}

@Article{Dasanayake2025,
AUTHOR = {Dasanayake, Pubudu Suranga and Baranauskas, Virginijus and Dervinis, Gintaras and Balasevicius, Leonas},
TITLE = {A Review of Mathematical Models in Robotics},
JOURNAL = {Applied Sciences},
VOLUME = {15},
YEAR = {2025},
NUMBER = {14},
ARTICLE-NUMBER = {8093},
ISSN = {2076-3417},

}

@article{iotti2025omniquad,
  title={OmniQuad: A wheeled-legged hybrid robot with omnidirectional wheels},
  author={Iotti, Francesco and Ranjan, Alok and Angelini, Franco and Garabini, Manolo},
  journal={Mechanism and Machine Theory},
  volume={214},
  pages={106125},
  year={2025},
  publisher={Elsevier}
}

\appendix

\end{document}